\documentclass[letterpaper, 10 pt, conference]{ieeeconf}  % Comment this line out
\IEEEoverridecommandlockouts                              % This command is only
\usepackage[para,online,flushleft]{threeparttable}
\usepackage{algorithm}
\usepackage{textcomp}
\usepackage[noend]{algorithmic}
\usepackage{booktabs}
\usepackage{multirow}
\usepackage{adjustbox}
\usepackage{graphicx}
\usepackage{siunitx}
\usepackage[utf8]{inputenc}
\usepackage[T1]{fontenc}
\usepackage{amsfonts}

\usepackage{amsthm}

\newtheorem*{remark}{Remark}
\usepackage{epsfig} % for postscript graphics files
\usepackage{amsmath} % assumes amsmath package installed
\usepackage{amssymb}  % assumes amsmath package installed
\usepackage{siunitx}
\usepackage{subcaption}
\usepackage{caption}
\usepackage{color}
\usepackage{theoremref}
\usepackage[noadjust]{cite}
\usepackage{amsthm}
\newtheorem*{theorem*}{Theorem}

\newcommand{\ma}[1]{\textcolor{black}{#1}}
\newcommand{\xuan}[1]{\textcolor{black}{#1}}

\title{\LARGE\bf Designing Multi-Stage Coupled Convex Programming with\\Data-Driven McCormick Envelope Relaxations for Motion Planning}
\author{Xuan Lin$^{1*}$, Min Sung Ahn$^{1*}$, and Dennis Hong$^{1}$% <-this % stops a space
\thanks{$^{1}$X. Lin, M. S. Ahn, and D. Hong are with the Robotics and Mechanisms Laboratory, Department of Mechanical and Aerospace Engineering, University of California, Los Angeles, CA 90095, USA.
        {\tt\small \{maynight,aminsung,dennishong\}@ucla.edu}}
\thanks{*X. Lin and M. S. Ahn contributed equally to this work.}
}

\begin{document}
\maketitle
\thispagestyle{empty}
\pagestyle{empty}
%%%%%%%%%%%%%%%%%%%%%%%%%%%%%%%%%%%%%%%%%%%%%%%%%%%%%%%%%%%%%%%%%%%%%%%%%%%%%%%%
%%note: must be manage risk in a "quantitative" way! Model and environmental uncertainty are coupled.
%% reduce to simpler deterministic optimization when I assume the noise follows gauusian

\begin{abstract}
% For multi-limbed robots motion planning with posture and force constraints tend to be a difficult optimization problem to solve due to variable coupling. As the problem size scales up, single stage solvers suffer from extended solving time, while typical decoupled multi-staged algorithms sequentially optimize each stage individually, resulting in infeasibility or local optimality. 
% We propose a multi-staged optimization framework with inter-stage coupling that allows the prior stages to preview following stages. By relaxing the bilinear constraints in the second stage into McCormick envelopes, they can be embedded into the first stage. 
% The envelopes are designed to ensure that solving the first stage guarantees a solution to the second stage. The planner is validated through multiple walking and vertical climbing tasks on a 10 kg hexapod robot. 
% Approximating the nonlinear constraints with smaller pieces of linear constraints usually results in extended solving time.

% of the nonlinear constraints that are tailored to the problem parameters. 

% We then input the learned constraints into the prior stages, allowing them to preview the following stages. 

For multi-limbed robots, motion planning with posture and force constraints tends to be a difficult optimization problem due to nonlinearities, which also present extended solve times. We propose a multi-stage optimization framework with data-driven inter-stage coupling constraints to address the nonlinearity. 
Both clustering and evolutionary approaches to find the McCormick envelope relaxations are used to find the problem-specific parameters. The learned constraints are then used in the prior stages, which provides advanced knowledge of the following stages. This leads to improved solve times and interpretability of the results. The planner is validated through multiple walking and %vertical 
climbing tasks on a 10 kg hexapod robot. 

\end{abstract}
% 
% 
% \begin{IEEEkeywords} Legged Robots, Motion and Path Planning,  Optimization and Optimal Control
%  \end{IEEEkeywords}

%%%%%%%%%%%%%%%%%%%%%%%%%%%%%%%%%%%%%%%%%%%%%%%%%%%%%%%%%%%%%%%%%%%%%%%%%%%%%%%%
\section{Introduction}
\label{Sec:introduction}
Legged robots present a unique advantage compared to their wheeled counterparts, where discrete contact with the environment can be made to maneuver discontinuous, complex terrains. 
To fully realize this capability,
% legged robots 
\ma{they}
often use motion planning to autonomously choose their footholds and plan their body movements to avoid slipping and tumbling. 
Optimization-based techniques have been exploited by researchers to resolve the motion planning problem 
% \cite{kuindersma2016optimization}\cite{winkler2018gait}\cite{lin2019optimization}\cite{ahn2018stable}. 
\ma{\cite{kuindersma2016optimization, winkler2018gait, lin2019optimization, ahn2018stable}.} 
The easiest way to implement optimization for motion planning would be to include all the linear or nonlinear equations into a single optimization problem. 
% Unfortunately, such an approach often suffers from time complexity \cite{winkler2018gait}. 
\ma{Unfortunately, such an approach often suffers from time complexity \cite{winkler2018gait}, although recent efforts to parallelize optimizations have shown speedups \cite{plancher2019realtime}.}
To reduce the solving time \ma{without such techniques}, one approach is to separate the entire problem into \ma{multiple} stages, where each stage only solves \ma{for} a 
% certain 
subset of \ma{the decision} variables 
% \cite{vernaza2009search}\cite{boyd2004convex}\cite{kalakrishnan2011learning}\cite{zucker2010optimization}. 
\ma{\cite{vernaza2009search, boyd2004convex, kalakrishnan2011learning, zucker2010optimization}.}
\ma{However, if each stage independently finds solutions without considering the subsequent stages' constraints, the later stages may result in no feasible solution.}
% However, issues can happen if each stage works independently without information from other stages. When the first stage is solving its subset from the entire problem, it only finds the optimal solution considering constraints on its own. This could leave the second stage, which receives results from the first stage, with no feasible solution. 
% This issue will be exacerbated if the problem scales up to more stages.
\ma{This issue exacerbates when the problem scales to many stages.}

\begin{figure}[!t]
		\centering
		\includegraphics[scale=0.40]{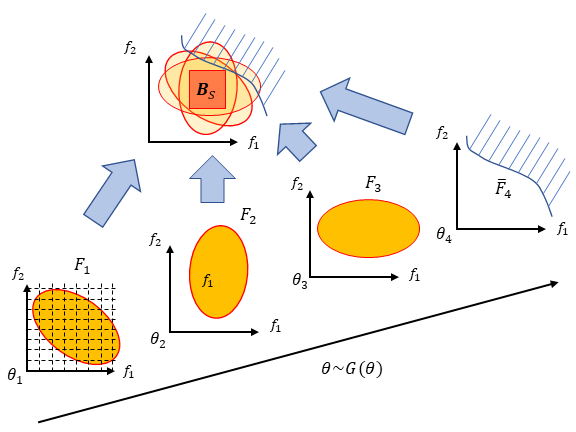}
		\caption {The proposed approach to find $\textbf{B}_{\mathcal{F}}$ inside the intersection of solution set $F_{i}$'s and exclude infeasible solution set $\overline{F}_{i}$'s. While we can finely grid the variable space and accurately approximate $F_{i}$'s in each grid (shown in $F_{1}$ plot), the proposed approach results in less integer variables.}
		\label{Fig:Figure_1}
\end{figure}

This paper proposes a multi-staged trajectory optimization (TO) formulation with inter-stage couplings
% that are 
learned from data. 
We relax the nonlinear constraints in the later stages into convex forms, and embed them into the prior stages, 
% allowing earlier stages to preview the information of later stages. 
% As a result, the first stage solves an approximated form of the original problem, and subsequent stages project the solutions back to the nonlinear manifolds to correct the error. 
\ma{resulting in the first stage solving an approximated form of the original problem and subsequent stages projecting the solutions back to the nonlinear manifolds to correct the error. }
While dealing with the TO problem as a whole is complicated, the clear interpretability of the convex optimization permits us to segment the constraints out, perform dedicated learning, and place them back into the structure. 
Learning is done using both clustering and genetic algorithm (GA), where we identify the regions that the data populate and fit constraints around that region. 
% The resultant formulation is tailored to the type of problem, thus the solving speed and the interpretability are improved.
The resultant formulation is tailored to the given type of problem, and solve time and interpretability are improved. 

While the proposed method is applicable for general constraints, we tackle \ma{a simple}
% one of the basic 
nonlinear constraint---the bilinear constraint---by embedding it as a McCormick envelope in the first optimization stage. McCormick envelopes are the best linear relaxation for bilinear constraints \cite{mccormick1976computability}. They are widely used to formulate bilinear constraints into mixed-integer convex programming (MICPs) \cite{dai2019global}, where the complete bilinear surface is divided evenly into multiple envelopes. 
This way of segmentation is general but usually results in too many integer variables and extended solving time. 
% Previous studies are conducted that learn optimal strategies with learning agents (e.g. neural-networks) to provide the best choices of integer variables to the MICP solver, reducing the solving time 
% \cite{bertsimas2020voice}\cite{cauligi2020learning}. 
\ma{To reduce solve time, previous studies learned optimal strategies with neural networks to provide the best choice of integer variables to the MICP solver \cite{bertsimas2020voice,cauligi2020learning}. }
However, since the number of strategies is an exponential function of the number of integer variables, the number of class labels quickly becomes intractable as the problem gets more complicated. 
It may be more efficient to directly conduct learning on the MICP framework to reduce the number of integer variables before it is handed to any learning agent. 
Instead of evenly dividing the bilinear surface into envelopes, we propose to adapt the placement of envelopes to the given type of problems, resulting in a reduced number of integer variables. Adaptive envelope tuning can be seen in mixed-integer nonlinear programming (MINLP)
% \cite{gleixner2013learning}\cite{nagarajan2016tightening}\cite{nagarajan2019adaptive} 
\ma{\cite{gleixner2013learning,nagarajan2016tightening,nagarajan2019adaptive}}
where the general idea is to fit envelopes to the more promising regions on the solution manifold as the problem is being solved. 
Our method makes use of learning to perform adaptation before the solving process is started. In this paper, we focus on solving the motion planning problem applied to multi-legged robot locomotion on multiple terrains. Hardware testings are done to show feasibility. Our contributions are:
\begin{enumerate}
\item A multi-stage coupled optimization-based motion planning framework with the later stages relaxed and embedded into the earlier stages.
\item Learning best McCormick envelope relaxation with clustering and GA techniques from data.
\item Demonstration of the planning results on hardware.
\end{enumerate}

\section{Problem setup}
\label{Sec:problem_setup}
%Should I mention that this constraint is equivalent to the whole body stiffness matrix in previous paper? - NO
% This section formulates the motion planning problem.
This section presents the formulations of the motion planning problem which is to be ultimately solved.

\begin{figure}[!t]
		\centering
		\includegraphics[scale=0.30]{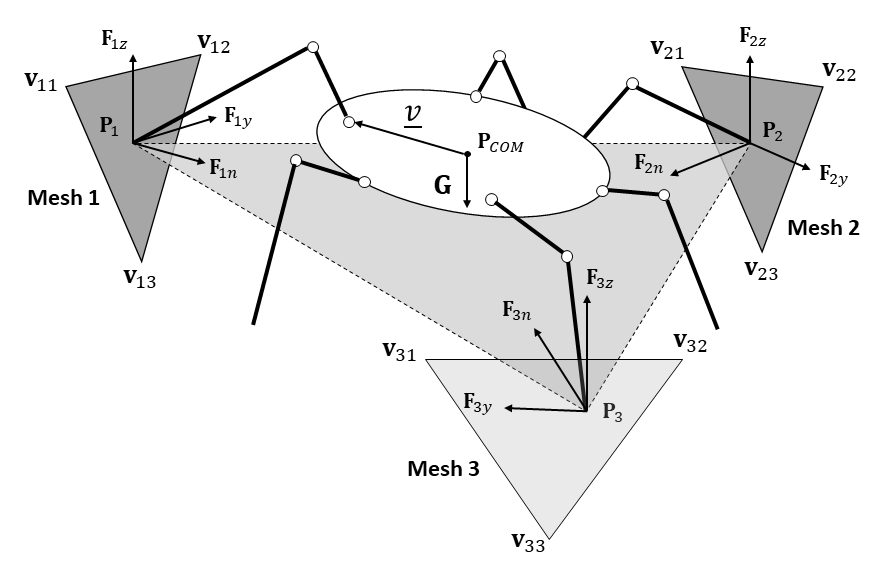}
		\caption {A robot making 3 contacts with the environment, subject to gravity. Contact planes are represented by triangular meshes with vertices $\textbf{v}_{sj} (i,j=1,2,3)$. $\textbf{F}_{sn}$, $\textbf{F}_{sy}$, $\textbf{F}_{sz}$ are constant vectors along the normal and two pre-defined shear directions. The three contact points form a contact triangle.}
		\label{Fig:notations}
\end{figure}

A multi-limbed robot is assumed to make $N$ point contacts (i.e. pure contact force, no contact moment) with the environment. We denote the contact points with index $i$ where $i=1, ..., N$. We model the environment with polygon meshes (
% \xuan{as an example,} 
\ma{e.g.} a triangular mesh is depicted in Fig. \ref{Fig:notations}). 
% With a perception system (e.g. an RGB-D camera), the vertices of the meshes $i$ (corresponding to limb $i$, denoted by $\textbf{v}_{ij}$ where $j$ is the index of vertices), and the normal direction $\textbf{n}_{i}$ is known to the robot. 
%Additionally, we assume that the vertices of the meshes $i$ (corresponding to the limb $i$ in contact with the mesh) are denoted by $\textbf{v}_{ij}$, where $j$ is the indices of the vertices, and the normal direction $\textbf{n}_{i}$ is known via perception system. 
\xuan{Additionally, we assume that  
% the vertices of the meshes $i$
\ma{mesh $i$'s vertices}
(corresponding to the limb $i$ in contact with the mesh) are denoted by $\textbf{v}_{iu}$, where 
$u$ is the indices of the vertices, $u=1, ..., U$ and $U$ is the number of vertices for mesh $i$.
% $u=1, ..., U$ is the indices of the vertices and $U$ is the number of vertices for mesh $i$.
The normal direction $\textbf{n}_{i}$ can be retrieved via \ma{a} perception system. If we define $\textbf{p}^{w}_{i}$ to be the position of toe $i$ within mesh $i$ with respect to the 
\ma{world frame:}
% origin of global coordinate system, we have:
}

\xuan{\begin{equation}
    \textbf{p}^{w}_{i} = \sum_{u}p^{w}_{iu}\textbf{v}_{iu}, \ \sum_{j}p^{w}_{iu}=1, \ p^{w}_{iu}\in[0,1]
\label{Eqn:position_mesh}
\end{equation}}

\xuan{If we define $\textbf{p}^{b}_{i}$ to be the position of toe $i$ with respect to the origin of \ma{the} body coordinate system, we have:}

\xuan{\begin{equation}
    \textbf{p}^{w}_{i} = \textbf{p}_{COM} +  \textbf{p}^{b}_{i}
\label{Eqn:p_world_2_p_body}
\end{equation}
where $\textbf{p}_{COM}$ is the position of the center of mass \ma{(COM)} with respect to the origin of the global coordinate system.}

% Consequently, for a contact point that lies within mesh $i$, the toe position $\textbf{p}_{i}$ is:

% \begin{equation}
%     \textbf{p}_{i} = \sum_{j}p_{ij}\textbf{v}_{ij}, \ \sum_{j}p_{ij}=1, \ p_{ij}\in[0,1]
% \label{Eqn:position_non-dim}
% \end{equation}

Let the contact force on limb $i$ be denoted by $\textbf{f}_{i}$. 
% Since the normal direction of mesh $i$ is known as $\textbf{n}_{i}$, we can define two mutually perpendicular shear directions to be $\textbf{y}_{i}$ and $\textbf{z}_{i}$ (they can be pre-defined for every mesh).%as long as it is independent of a specific footstep plan!
Since $\textbf{n}_{i}$ is known, we can define two mutually perpendicular shear directions $\textbf{y}_{i}$ and $\textbf{z}_{i}$, which can be pre-defined for every mesh. 
The three directional vectors form a local mesh coordinate frame, 
% % The contact force thus can be decomposed into three components
% and the contact force can be decomposed as $\textbf{f}_{in}$, $\textbf{f}_{iy}$, $\textbf{f}_{iz}$ in the mesh frame. 
% $\textbf{f}_{in}$ the normal direction force, while $\textbf{f}_{iy}$ and $\textbf{f}_{iz}$ represent the two shear forces components. 
% Thus $\textbf{f}_{i} = \textbf{f}_{in} + \textbf{f}_{iy} + \textbf{f}_{iz}$. 
\ma{and the contact force can be represented as $\textbf{f}_{i} = \textbf{f}_{in} + \textbf{f}_{iy} + \textbf{f}_{iz}$ in the mesh frame, where $\textbf{f}_{in}$ is the normal force, and $\textbf{f}_{iy}$ and $\textbf{f}_{iz}$ are the shear force components.}
Define $\textbf{F}_{in}$, $\textbf{F}_{iy}$, $\textbf{F}_{iz}$ as constant vectors along the normal and two shear directions on mesh $i$, we can express the dimensional force components using the non-dimensional force components $f_{in}$, $f_{iy}$, $f_{iz}$ as $\textbf{f}_{ik} = f_{ik}\textbf{F}_{ik}, \;\; k=n,y,z$. The total contact force is:
% Let the robot gravity vector be $\textbf{G}$. We define $\textbf{F}_{in}$, $\textbf{F}_{iy}$, $\textbf{F}_{iz}$ to be constant vectors along the normal and two pre-defined shear directions on mesh $i$, all with norm $|\textbf{F}|=1.5G$, and express the dimensional force components by non-dimensional force components $f_{in}$, $f_{iy}$, $f_{iz}$:

\begin{equation}
    \textbf{f}_{i} = \sum_{k=n,y,z}f_{ik}\textbf{F}_{ik}
\label{Eqn:force_non-dim}
\end{equation}

\xuan{Similarly, defining $\textbf{P}_{x}$, $\textbf{P}_{y}$, $\textbf{P}_{z}$ as constant vectors along the 3 axes of the 
% global coordinate system
\ma{world frame,}
$\textbf{p}^{b}_{i}$ can be expressed as:}

\xuan{\begin{equation}
    \textbf{p}^{b}_{i} = \sum_{j=x,y,z}p^{b}_{ij}\textbf{P}_{j}
\label{Eqn:position_non-dim}
\end{equation}
Both position and force characteristic quantities are chosen such that $p^{b}_{ij} \in [-1, 1]$, $f_{ik} \in [-1, 1]$.}

\xuan{Also note that equation (\ref{Eqn:force_non-dim}) and (\ref{Eqn:position_non-dim}) non-dimensionalized the force and position variables $\textbf{f}_{i}$ and $\textbf{p}^{b}_{i}$ into
% non-dimensinoal 
quantities $f_{ik}$ and $p^{b}_{ij}$.} %Bear in mind that it is suggested to non-dimensionalize the variables 
This is suggested before utilizing certain mathematical operations to avoid unit mismatch as seen in \cite{ponton2016convex}.
% (TODO: Why not say this before nondimensionalizing? - \xuan{I have moved the previous section to here})

Having 
% setup 
\ma{set up} the contact positions and forces, whole body constraints can be imposed. 
% Having set up the contact positions and forces, we are now ready to impose robot whole body constraints.
The robot motion is assumed to be quasi-static, 
thus the robot is always subject to the static equilibrium constraint:
% \ma{thus the robot is in static equilibrium:}

\begin{equation}
\sum_{i=1}^{N} \textbf{f}_{i} + \textbf{F}=0
\label{Eqn:Static_equilibrium_force}
\end{equation}

% \begin{equation}
% \sum_{i=1}^{N} \textbf{p}_{i}\times\textbf{f}_{i}+\textbf{M}+\textbf{p}_{COM}\times\textbf{F}=0
% \label{Eqn:Static_equilibrium_moment_2}
% \end{equation}

\begin{equation}
\sum_{i=1}^{N} \textbf{p}^{b}_{i}\times\textbf{f}_{i}+\textbf{M}=0
\label{Eqn:Static_equilibrium_moment_2}
\end{equation}
where $\textbf{F}$ and $\textbf{M}$ are known external forces (gravity $\textbf{G}$ in this work) and moments. 
%and $\textbf{p}_{COM}$ is the position of the center of mass. 
% It is also assumed that the COM is always at the geometric center, irrespective of the robot limb motion.
\ma{COM is also assumed to always be at the geometric center, irrespective of the robot limb motion.}

\xuan{Equation (\ref{Eqn:Static_equilibrium_moment_2}) introduces a bilinear term $\textbf{p}^{b}_{i}\times\textbf{f}_{i}$. 
Utilizing (\ref{Eqn:force_non-dim}) and (\ref{Eqn:position_non-dim}), the bilinear term can further be expressed as:}

\xuan{\begin{equation}
    \textbf{p}^{b}_{i}\times\textbf{f}_{i} = \sum_{j=x,y,z} \sum_{k=n,y,z} p^{b}_{ij}f_{ik}\textbf{P}_{j}\times\textbf{F}_{ik}
\label{Eqn:Static_equilibrium_moment_3}
\end{equation}}

\xuan{To isolate the bilinear terms, let the moment variables be:}

\xuan{\begin{equation}
    m_{ijk} = p^{b}_{ij}f_{ik}
\label{Eqn:change_of_variable}
\end{equation}
where $m_{ijk}$'s are the moment components. Plugging (\ref{Eqn:change_of_variable}) into (\ref{Eqn:Static_equilibrium_moment_3}), and further back into (\ref{Eqn:Static_equilibrium_moment_2}) results in:}

\xuan{\begin{equation}
    \sum_{i=1}^{N} \sum_{j=x,y,z} \sum_{k=n,y,z} m_{ijk}\textbf{P}_{j}\times\textbf{F}_{ik} + \textbf{M} =0
\label{Eqn:Static_equilibrium_moment_4}
\end{equation}}

% In (\ref{Eqn:Static_equilibrium_moment_2}), except for $\textbf{p}_{i}\times\textbf{f}_{i}$ terms which are bilinear in the decision variables, the rest is linear. 
% Utilizing (\ref{Eqn:position_non-dim}) and (\ref{Eqn:force_non-dim}), the bilinear term can further be expressed as:

% \begin{equation}
%     \textbf{p}_{i}\times\textbf{f}_{i} = \sum_{j=1}^{N} \sum_{k=n,y,z} p_{ij}f_{ik}\textbf{v}_{ij}\times\textbf{F}_{ik}
% \label{Eqn:Static_equilibrium_moment_3}
% \end{equation}

% To isolate the bilinear terms, let the moment variables be:

% \begin{equation}
%     m_{ijk} = p_{ij}f_{ik}
% \label{Eqn:change_of_variable}
% \end{equation}
% where $m_{ijk}$s are the moment components. Plugging (\ref{Eqn:change_of_variable}) into (\ref{Eqn:Static_equilibrium_moment_3}), and further back into (\ref{Eqn:Static_equilibrium_moment_2}) results in:

% \begin{equation}
%     \sum_{i=1}^{N} \sum_{j=1}^{N} \sum_{k=n,y,z} m_{ijk}\textbf{v}_{ij}\times\textbf{F}_{ik} + \textbf{M} + \textbf{p}_{COM}\times\textbf{F}=0
% \label{Eqn:Static_equilibrium_moment_4}
% \end{equation}

% Index s has some issue.

% Note that the cross products $\textbf{v}_{ij}\times\textbf{F}_{ik}$ are known quantities, given $\textbf{v}_{ij}$ and $\textbf{F}_{ik}$ retrieved from perception systems.

In the predominant case where legged robots are not equipped with grippers on its toes, the point contacts with the environment are pure frictional contacts.
% In this paper, the multi-limbed robot is not equipped with grippers on its toe, so the point contacts on the environment are pure frictional contacts.
When the robot places its toes, the contact forces are subject to friction cone constraints. 
Given (\ref{Eqn:force_non-dim}) and defining $\mu$ as the friction coefficient, the constraint can be written as:
% Let the coefficient of friction be $\mu$. Given equation (\ref{Eqn:force_non-dim}), the constraint can be written as:

\begin{equation}
    \sqrt{f_{iy}^{2}+f_{iz}^{2}} < {\mu}f_{in}
\label{Eqn:friction_cone}
\end{equation}

To plan a complete quasi-static motion to the specified goal, the motion planner generates a series of ``key frames"---body COM positions, body orientations, and footstep positions---to the goal position. The number of key frames is pre-specified as $M$, and each key frame posture needs to satisfy kinematics constraints. Between two consecutive key frames, step size constraints are enforced. Similar to \cite{lin2019optimization}, the following constraints are used:
\begin{equation}
\begin{aligned}
&&&\text{for $r = 1, \ldots, \textit{M}, $}\\
&&&\Delta\textbf{\underline{p}}_{min}
 \le \| \textbf{\underline{p}}_{COM}[r]-\textbf{\underline{p}}_{COM}[r-1] \|_{2} \le \Delta\textbf{\underline{p}}_{max} \\
&&&\Delta\textbf{\underline{P}}_{min} \le \| \textbf{\underline{p}}_{i}[r]-\textbf{\underline{p}}_{i}[r-1] \|_2 \le \Delta\textbf{\underline{P}}_{max} \\
&&&\Delta\underline{\Theta}_{min} \le \| \underline{\Theta}_{b}[r]-\underline{\Theta}_{b}[r-1] \|_2 \le \Delta\underline{\Theta}_{max} \\
&&&\| \textbf{\underline{p}}_{i}[r]-\textbf{\underline{p}}_{COM}[r]-\textbf{R}[r]\underline{v}\|_2 \le\Delta_{FK} \\
\medskip
\label{Eqn:kinematics_constraints}
\end{aligned}
\end{equation}
where ${\Theta}_{b}[r]$ is body orientation, $\Delta\textbf{\underline{p}}_{min}$, $\Delta\textbf{\underline{p}}_{max}$, $\Delta\textbf{\underline{P}}_{min}$, $\Delta\textbf{\underline{P}}_{max}$, $\Delta\underline{\Theta}_{min}$, \ma{and} $\Delta\underline{\Theta}_{max}$ are bounds for the toe, body COM, and orientation step sizes. The limb workspace is simplified into a sphere, with $\Delta_{FK} \in \mathbb{R}$ its radius. $\underline{v}$ is the constant shoulder vector from body COM to the first joint of the limb (depicted in Fig. \ref{Fig:notations}). $\textbf{R}[r]$ is the rotation matrix as a function of $\underline{\Theta}_{b}[r]$. We assume the body rotation angles are small ($<$ 15 degrees), thus the rotation matrix can be linearized in terms of $\underline{\Theta}_{b}[r] = [\alpha, \beta, \gamma]$ \cite{lin2019optimization}.

% \begin{equation}
% \textbf{R}(\underline{\Theta}_{b}[r]) =
% \begin{bmatrix}
% 1      &  -\gamma   &   \beta   \\
% \gamma   &     1     &   -\alpha  \\
% -\beta  &    \alpha  &     1     \\
% \end{bmatrix}
% \label{Eqn:linear_rotation}
% \end{equation}

In summary, \xuan{there are kinematics constraints (\ref{Eqn:kinematics_constraints}) (convex), (\ref{Eqn:position_mesh}) (\ref{Eqn:p_world_2_p_body}) (\ref{Eqn:position_non-dim}) (linear), static equilibrium constraints (\ref{Eqn:force_non-dim}) (\ref{Eqn:Static_equilibrium_force}) (\ref{Eqn:Static_equilibrium_moment_4}) (linear), with additional constraints (\ref{Eqn:change_of_variable}) (bilinear), and friction cone constraints (\ref{Eqn:friction_cone}) (convex).} The objective function minimizes the distance of the planned final configuration to the goal configuration and penalizes the step size in similar fashion as \cite{kuindersma2016optimization}\cite{lin2019optimization}. Let us denote the complete problem by $P$. This problem is parametrized by $\theta$ which is the terrain geometry \xuan{$\textbf{v}_{iu}$}. Fig. \ref{Fig:notations} illustrates the scene and the notations that are used in this paper. For simplicity, the decision variables are grouped into two sets - kinematics variables $\mathit{\Gamma}_{p} = \{ \ \textbf{\underline{p}}_{i}[r], \ p_{ij}[r], \ \textbf{\underline{p}}_{COM}[r], \ \underline{\Theta}_{b}[r] \ \}$ and force variables $\mathit{\Gamma}_{f} = \{ \ \textbf{f}_{i}[r], \ f_{ik}[r], \ m_{ijk}[r] \ \}$.

% In this paper, we treat the bilinear constraints with a 2-stage relax-project method 

\begin{figure}[!t]
		\centering
		\includegraphics[scale=0.52]{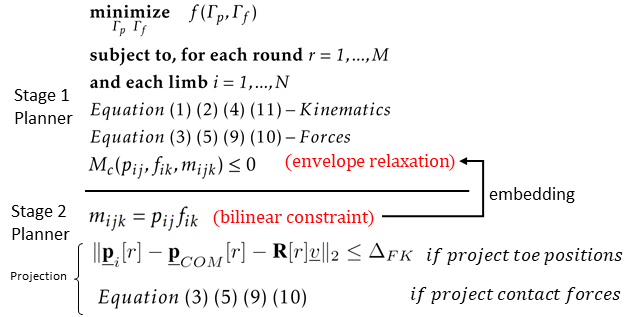}
		\caption {Optimization formulation of the motion planning problem}
		\label{Fig:complete_formulation}
\end{figure}

A standard approach to convert bilinear constraints into MICPs is to grid the variable space (shown in Fig. \ref{Fig:Figure_1} with $F_{1}$) and use McCormick envelopes to approximate the nonlinear constraint inside each grid. This approach accurately describes solutions for any problem parameter $\theta$, 
% but usually result in binary variables of hundreds 
\ma{but introduces hundreds of binary variables into the optimization problem}
\cite{dai2019global}, hence solving speed is slow. In this paper, we solve problem $P$ by introducing a 2-stage convex optimization process $P_{1} \rightarrow P_{2}$.
%, with McCormick envelopes in $P_{1}$ learned from data. 
% At the first stage $P_{1}$, we approximate the bilinear constraints using McCormick envelopes. The first stage solves both sets of bilinear variables $p_{ij}$ and $f_{ik}$ approximately. 
\ma{During the first stage $P_{1}$, we approximate the bilinear constraints in $P_{2}$ using McCormick envelopes and solve for both sets of bilinear variables $p_{ij}$ and $f_{ik}$.}
% At the 2nd
\ma{In the second} stage $P_{2}$, we choose to keep one set of bilinear variables in the solution of $P_{1}$ and project the other set onto the bilinear surface. 
If one set of bilinear variables are given,  
%becomes linear, 
\ma{the \xuan{bilinear constraint (\ref{Eqn:change_of_variable})} in $P_{2}$ is linearized.}
% constraint (\ref{Eqn:change_of_variable}) in $P_{2}$ is linearized. 
Because of the McCormick envelope in 
% the 1st stage
\ma{$P_1$}, the 2-stage planner is \textit{coupled}, 
% meaning we embed an approximation of the nonlinear constraint carrying information of the 2nd stage into the 1st stage. 
with the approximation of the nonlinear constraint in 
% the 2nd stage 
\ma{$P_2$}
embedded in
% the 1st stage
\ma{$P_1$}. 
%The 2-stage planner with embedding is shown in Fig. \ref{Fig:complete_formulation}.
This process is shown in Fig. \ref{Fig:complete_formulation}.

We propose to learn the best envelope to a certain type of problem, such that the envelope formulation captures the heuristics behind the given type of input. Since trajectory optimization solves a series of postures with identical mathematical formulation, it is sufficient to learn the constraints on a single posture. We generate data by solving the problem with $M=1$ using the accurate MICP formulation \cite{dai2019global}, and collect data for bilinear variables. We then fit McCormick envelopes around the regions where data clusters, excluding the infeasible regions. Through this approach, the relaxation can be made tighter with a fixed number of envelopes.

\section{Learning setup} 
\label{Sec:mpdesign}

We present the mathematical setup for the learning problem. 
% The solutions for similar types of problems could be densely populated in a certain region. 
% By using learning methods, we could identify those regions and form a ``tighter'' McCormick envelope, eliminating regions in the variables' space where infeasible solutions are expected. 
\ma{As solutions of similar types of problems could be densely populated in certain regions, learning methods could be used to identify those regions to form a ``tighter" McCormick envelope. This eliminates regions in the variables' space where infeasible solutions are expected.}
% The idea of learning is to identify those overlapping feasible regions and construct envelopes around them. % to construct the envelopes.

% The McCormick envelope \cite{castro2015tightening} gives a best linear approximation for the bilinear constraint parametrized by 4 variables representative of the envelope bounds [$p^{L}$, $p^{U}$], [$f^{L}$, $f^{U}$]. 
% Fig. (\ref{Fig:2D_envelope}) shows a sectional view of a bilinear surface with a pre-defined $f$ bounded in [-1, 1], which results in the largest envelope relaxation. 
% The crux of this contribution is to investigate how much ``tighter'' the bounds could be made to obtain a better envelope relaxation. 
% Accurate bounds can be vital in the feasibility of the optimization problem; when the envelope is simply at its largest, infeasible solutions become feasible candidates. 
% As the envelope shrinks, less infeasible solutions will be deemed feasible. %as shown in the yellow region. 
% However, this occurs at the cost of feasible solutions becoming infeasible. %(e.g. the blue line outside the shrunk envelope). 

% Mathematical analysis shows that optimal bounds for the envelope do exist.
\xuan{Our algorithm \ma{seeks} the overlapping regions for bilinear variables $(\textbf{p}, \textbf{f})$. The position variable $\textbf{p}$ to be learned is $\textbf{p}^{b}$. We omit the superscript for simplicity. Suppose we define problem $P(\theta_{i})$ where $\theta_{i}$ is the parameter drawn from a distribution $G(\theta)$ and represents the terrain shape in our work. If $P(\theta_{i})$ is feasible, we define the \textit{solution set} $X(\theta_{i}) = \{\textbf{x}|\textbf{x}=(\textbf{p}, \textbf{f}, \textbf{m}, \textbf{y}) \ \text{feasible for} \ P(\theta_{i})\}$. $\textbf{p}, \textbf{f}, \textbf{m}$ are the variables that satisfy the bilinear constraint $m_{d}=p_{d}f_{d}$ in each dimension $d=1, ..., D$, while $\textbf{y}$'s are other optimization variables. If $P(\theta_{i})$ is infeasible, we define the \textit{infeasible solution set} $\overline{X}(\theta_{i})$ by removing the bilinear constraint from $P(\theta_{i})$ that generates $\overline{P}(\theta_{i})$, and $\overline{X}(\theta_{i}) = \{\textbf{x}|\textbf{x}=(\textbf{p}, \textbf{f}, \textbf{m}, \textbf{y}) \ \text{feasible for} \ \overline{P}(\theta_{i})\}$. A McCormick envelope relaxation of $m_{d}=p_{d}f_{d}$ \cite{castro2015tightening} can be defined tightly over a pair of lower/upper bounds $[\textbf{p}_{d}^{L}, \textbf{p}_{d}^{U}]$ and $[\textbf{f}_{d}^{L}, \textbf{f}_{d}^{U}]$
%. For a given set of vector bounds $\textbf{p}_{d}^{L}$, $\textbf{p}_{d}^{U}$, $\textbf{f}_{d}^{L}$ and $\textbf{f}_{d}^{U}$ 
that describes a rectangular region $\textbf{B}$ over $[\textbf{p}^{T}, \textbf{f}^{T}]^{T}$. McCormick envelopes $M_{c}(\textbf{B})$ satisfy the property such that $[\textbf{p}^{T}, \textbf{f}^{T}]^{T} \in \textbf{B}$ and $m_{d}=p_{d}f_{d}, \ \forall d$ implies $M_{c}(\textbf{B}) \leq 0$. Furthermore, $[\textbf{p}^{T}, \textbf{f}^{T}]^{T} \notin \textbf{B} \Rightarrow M_{c}(\textbf{B}) \nleq 0$.} %\xuan{(this does not mean $M_{c}(\textbf{B}) \geq 0$ as this is a vector constraint)}.} 
\xuan{Define $S_{i}$ as the projection of $X(\theta_{i})$ onto $\textbf{p}\times\textbf{f}$ (``$\times$'' is Cartesian product) subspace giving all feasible $[\textbf{p}^{T}, \textbf{f}^{T}]^{T}$ for problem $P(\theta_{i})$. Then, $\forall (\textbf{p}, \textbf{f}) \in S_{i}$, $\exists$ $(\textbf{m}, \textbf{y})$ such that $(\textbf{p}, \textbf{f}, \textbf{m}, \textbf{y})$ is feasible for $P(\theta_{i})$. We also define the projection of \textit{infeasible solution set} of $[\textbf{p}^{T}, \textbf{f}^{T}]^{T}$ as $\overline{S}_{i}$ by projecting $\overline{X}(\theta_{i})$ onto $\textbf{p} \times \textbf{f}$ subspace.} 

\xuan{Assume that  for any feasible $P(\theta_{i})$ that generates $S_{i}$, there exists an overlapping region: $\mathcal{S} = \cap S_{i} \neq \emptyset$. We can find a rectangular region $\textbf{B}_{\mathcal{S}}$ inside $\mathcal{S}$ to construct a McCormick envelope relaxation $M_{c}(\textbf{B}_\mathcal{S}) \leq 0$, and define the 2-step optimization process $P_{1} \rightarrow P_{2}$. $P_{1}$ is defined by replacing the biliner constraints in $P(\theta_{i})$ with the McCormick relaxation, and gives a solution $(\widetilde{\textbf{p}}, \widetilde{\textbf{f}}, \widetilde{\textbf{m}}, \widetilde{\textbf{y}})$. 
$P_{2}$ has identical constraints as $P$, but
% keeps
\ma{uses} 
$(\widetilde{\textbf{f}}, \widetilde{\textbf{y}})$ (or $(\widetilde{\textbf{p}}, \widetilde{\textbf{y}})$) from $P_{1}$
% and solves for 
\ma{to solve for}
$(\textbf{p}^{*}, \textbf{m}^{*})$ (or $(\textbf{f}^{*}, \textbf{m}^{*})$), 
% such that 
\ma{as}
the bilinear constraints are linearized.
}
\ma{
\begin{theorem*}
If \xuan{there exist a McCormick envelope $M_{c}(\textbf{B}_\mathcal{S}) \leq 0$ that satisfies} the following 3 conditions:
\begin{enumerate}
    \item \xuan{$\forall \ \theta_{i}$ that makes $P(\theta_{i})$ feasible and generates $S_{i}$, $\textbf{B}_{\mathcal{S}} \cap S_{i} \neq \emptyset$.} 
    \item $\forall \ \theta_{i}$ that makes $P(\theta_{i})$ infeasible, $\textbf{B}_{\mathcal{S}}\cap \overline{S}_{i}=\emptyset$. 
    \item (\textit{Complete Projectability}) 
    %All \xuan{$(\textbf{f}, \textbf{y})$ (or $(\textbf{p}, \textbf{y})$)} associated with $\textbf{B}_\mathcal{S}$ for $P_{1}$ are feasible for $P$ (
    $\forall \textbf{f} \in \textbf{B}_{\mathcal{S}}$, $\forall \textbf{y}$, \xuan{(or $\forall \textbf{p} \in \textbf{B}_{\mathcal{S}}$, $\forall \textbf{y}$)} such that $(\textbf{f}, \textbf{y})$ \xuan{(or $(\textbf{p}, \textbf{y})$)} is feasible for $P_{1}$, \xuan{$(\textbf{f}, \textbf{y})$ (or $(\textbf{p}, \textbf{y})$)} is also feasible for $P$. 
\end{enumerate}
\xuan{Then the 2-step process $P_{1} \rightarrow P_{2}$ defined above satisfy:}
\begin{enumerate}
    \item \xuan{$P$ is feasible $\Rightarrow$ $P_{1} \xrightarrow{} P_{2}$ is feasible. In addition, any feasible solution for $P_{1} \xrightarrow{} P_{2}$ is also feasible for $P$.}
    \item \xuan{$P$ is infeasible $\Rightarrow$ $P_{1}$ is infeasible.}
\end{enumerate}
\end{theorem*}
\xuan{Note that depending on how $P_{2}$ projects the solution of $P_{1}$, the required projectability condition (3) is different. If we keep $(\textbf{f}, \textbf{y})$ and project $(\textbf{p}, \textbf{m})$, then we need $(\textbf{f}, \textbf{y})$ to be feasible for $P$, and vise versa.}
\begin{proof}~ \\
% \begin{enumerate}
1) Suppose $P$ is feasible with solution subspace $S_{i}$. Since \xuan{$\textbf{B}_{\mathcal{S}} \cap S_{i} \neq \emptyset$}, $\exists [\textbf{p}_{i}^{T}, \textbf{f}_{i}^{T}]^{T} \in \xuan{\textbf{B}_{\mathcal{S}} \cap {S_{i}}}$. \xuan{$[\textbf{p}_{i}^{T}, \textbf{f}_{i}^{T}]^{T} \in S_{i} \Rightarrow$ } $\exists (\textbf{m}_{i}, \textbf{y}_{i})$ such that \xuan{$\textbf{x}=(\textbf{p}_{i}, \textbf{f}_{i}, \textbf{m}_{i}, \textbf{y}_{i})$} is feasible for $P$ \xuan{(satisfying each constraint except $M_{c}(\textbf{p}_{i}, \textbf{f}_{i}, \textbf{m}_{i}) \leq 0$)}. \xuan{In particular, $(\textbf{p}_{i}, \textbf{f}_{i}, \textbf{m}_{i})$ satisfies the bilinear constraint $m_{d}=p_{d}f_{d}$, $\forall d$. This together with $[\textbf{p}_{i}^{T}, \textbf{f}_{i}^{T}]^{T} \in \textbf{B}_{\mathcal{S}}$ implies $M_{c}(\textbf{p}_{i}, \textbf{f}_{i}, \textbf{m}_{i}) \leq 0$}. Thus $P$ is feasible $\Rightarrow P_{1}$ is feasible. \\
\xuan{Now we show that $P_{1}$ is feasible $\Rightarrow$ $P_{2}$ is feasible, and the solution of $P_{1} \xrightarrow{} P_{2}$ is feasible for $P$.
Any feasible solution $(\textbf{p}, \textbf{f}, \textbf{m}, \textbf{y})$ for $P_{1}$ satisfies $M_{c}(\textbf{p}, \textbf{f}, \textbf{m}) \leq 0 \Rightarrow (\textbf{p}, \textbf{f}) \in \textbf{B}_{\mathcal{S}}$. By condition (3), $(\textbf{f}, \textbf{y})$ (or $(\textbf{p}, \textbf{y})$) is feasible for $P$, which means $\exists (\widetilde{\textbf{p}}, \widetilde{\textbf{m}})$ (or $(\widetilde{\textbf{f}}, \widetilde{\textbf{m}})$) such that $\widetilde{\textbf{x}} = (\widetilde{\textbf{p}}, \textbf{f}, \widetilde{\textbf{m}}, \textbf{y})$ (or $(\textbf{p}, \widetilde{\textbf{f}}, \widetilde{\textbf{m}}, \textbf{y})$) satisfies $P$. Since $P_{2}$ has identical constraints as $P$, $\widetilde{\textbf{x}}$ is feasible for $P_{2}$. As $P_{2}$ is feasible and has identical constraints as $P$, any solution for $P_{2}$ is feasible for $P$.\\}
2) By condition (2), $\textbf{B}_{\mathcal{S}}\cap \overline{S}_{i}=\emptyset$, $\forall [\textbf{p}^{T}, \textbf{f}^{T}]^{T} \in \overline{S}_{i}$ (satisfies all but the bilinear constraints), $[\textbf{p}^{T}, \textbf{f}^{T}]^{T} \notin \textbf{B}_{\mathcal{S}}$, thus $M_c(\xuan{\textbf{p}}, \textbf{\textbf{f}}, \forall \textbf{m}) \nleq 0$ $\Rightarrow$ $P_{1}$ is infeasible.
% \end{enumerate}
\end{proof}
}

\begin{remark}
It may look like result 2) is too strong. Normally, a relaxation is feasible doesn't guarantee that the original problem is feasible. However, in this case it does. Basically, 2) shifts the region of relaxation on (\textbf{p}, \textbf{f}) completely away from the \textbf{danger zone} - region that potentially makes infeasible problems feasible. The relaxation will admit new \textbf{m} points. However, no matter what \textbf{m} is, since (\textbf{p}, \textbf{f}) is infeasible for the original problem, (\textbf{p}, \textbf{f}, \textbf{m}, \textbf{y}) wll not be feasible. A good example is that, for some problem, $\textbf{m}=\textbf{p}\textbf{f}$ does not constraint anything. This means whenever $\theta_{i}$ is infeasible for $P$, $\overline{P}$ is still infeasible. In this case, the envelope can be all the domain, as other constraints for $P$ are already making the problem infeasible. 
\end{remark}

The idea
% of
\ma{behind}
finding \xuan{$\textbf{B}_{\mathcal{S}}$} is shown in Fig. \ref{Fig:Figure_1}.
% (The issue here is you cannot project \textbf{y}. Basically, you should be looking for f and something else, otherwise there is no need to solve any optimization -> just find f)
Condition \xuan{(3)} guarantees projectability for any point in \xuan{$\textbf{B}_{\mathcal{S}}$}. Strictly satisfying it guarantees $P_{1}$ will not give solutions that cause $P_{2}$ to be infeasible. In practice, verifying condition \xuan{(3)} is very difficult. We put additional safety factors to make constraints in $P_{1}$ even tighter to reduce the the relaxation (at the risk of $P_{1}$ not a complete relaxation of $P$). \xuan{As an simple extension of this paper, one can also learn an overlapping region of $y_{i}$'s and formulate as an additional constraint into $P_{1}$. This guarantees that $P_{1}$ does not give any ``strange'' $y_{i}$ that makes $P_{2}$ infeasible.} Another approach is to set an optimal criteria for $P_{1}$ and guarantee that the optimal solution of $(\textbf{f}, \textbf{y})$ is always projectable to $P_2$
% (one example for optimal control is proven by \cite{acikmese2007convex}).
\ma{\cite{acikmese2007convex}.}
% One interesting future work is to learn an objective function that guarantees optimal projectability. If with all those efforts, the projection still fails, the problem is over-relaxed and the envelope should be broken into smaller ones.
If the projection still fails with the above efforts, the problem is over-relaxed, suggesting that the envelope should be divided into multiple smaller pieces. 
% However, interesting future work remains where potentially, an objective function that guarantees projectability at optimal points exist. 
\ma{Interesting future work remains where an objective function that guarantees projectability at optimal points could potentially exist.} \xuan{In addition, if we do not confine the formulation to convex ones, we can use nonlinear optimization (NLPs) in $P_{2}$ to do the projection. In this case, we only keep $y_{i}$ from $P_{1}$ and project $(\textbf{p}, \textbf{y})$ simultaneously. Properly initialized NLPs can have fast speed with a solvable rate close to 100\% \cite{dai2019global}.} \xuan{Finally,} even if \xuan{$\cap S_{i} = \emptyset$}, we can still identify multiple mutually exclusive \xuan{$\textbf{B}_{\mathcal{S}}$}'s that give multiple envelopes. An example is shown in the next section. 

\begin{algorithm}[t] 
\small 
\algsetup{linenosize=\small}
\caption{$\operatorname{DataGeneration}$}
\label{datacollection}
\textbf{Input} Number of samples $N$
\begin{algorithmic}[1] \label{alg:datagen}
\STATE Initialize feasible solution set \xuan{$S$} and infeasible solution set \xuan{$\overline{S}$}
\WHILE{$i$ < $N$}
\STATE Sample $\theta_{i} \sim G(\theta)$
\STATE Solve $P(\theta_{i})$ with high precision MICP
\IF{\xuan{($\textbf{p}_{i}$, $\textbf{f}_{i}$)} is a feasible solution for $P(\theta_{i})$}
\STATE Add \xuan{($\textbf{p}_{i}$, $\textbf{f}_{i}$)} to \xuan{$S$}
%\STATE Sample $n$ points \xuan{($\textbf{p}_{i}^{o}$, $\textbf{f}_{i}^{o}$)} around \xuan{($\textbf{p}_{i}$, $\textbf{f}_{i}$)}
  %\IF{$P(\theta_{i})$ is infeasible for \xuan{($\textbf{p}_{i}^{o}$, $\textbf{f}_{i}^{o}$)}}
  %\STATE Add \xuan{($\textbf{p}_{i}^{o}$, $\textbf{f}_{i}^{o}$)} to \xuan{$\overline{S}$}
  %\ENDIF
\ELSE
\STATE Remove bilinear constraint (\ref{Eqn:change_of_variable}) to produce problem $\overline{P}(\theta_{i})$
\STATE Solve $\overline{P}(\theta_{i})$
\IF{\xuan{($\overline{\textbf{p}}_{i}$, $\overline{\textbf{f}}_{i}$)} is a feasible solution for $\overline{P}(\theta_{i})$}
\STATE Add \xuan{($\overline{\textbf{p}}_{i}$, $\overline{\textbf{f}}_{i}$)} to \xuan{$\overline{S}$}
\ENDIF
\ENDIF
\STATE Increase $i$
\ENDWHILE
\RETURN \xuan{$S$}, \xuan{$\overline{S}$}
\end{algorithmic} 
\end{algorithm}

%The claim that we presented above gives two options: we can construct envelopes based on the overlapping region for $\textbf{p}$ or $\textbf{f}$. We choose to find the overlapping region for $\textbf{f}$. The reason is for the legged locomotion problems, $\textbf{f}$ is closely related to the terrain type. For walking, feasible $\textbf{f}$'s tend to have large normal forces, while for wall climbing, feasible $\textbf{f}$s tend to have both large normal and shear forces. Thus $\textbf{f}$s are expected to have distinct regions of feasibilities that can be learned. 
We present two data-based approaches to identify \xuan{$\textbf{B}_{\mathcal{S}}$}, one based on clustering to directly fit envelopes and the other one based on evolutionary algorithms.

\subsubsection{Clustering Approach}
Based on the high-accuracy MICP formulation \cite{dai2019global}, we can generate feasible solutions or prove infeasibility for problems $P(\theta_{i})$ sampled from $G(\theta)$. We recognize that if we sample the same amount of feasible solutions inside each \xuan{$S_{i}$}, the overlapping region \xuan{$\mathcal{S}$} will receive more samples, indicating a clustering approach may identify \xuan{$\mathcal{S}$}. We also need to draw samples from
%of infeasible solutions outside \xuan{$S_{i}$} as well as 
the \textit{infeasible solution set} \xuan{$\overline{S}_{i}$} and make the envelope exclude those points. The data generation algorithm is shown in Algorithm \ref{alg:datagen}, where we generate a random terrain from a pre-defined distribution and collect the subsequent optimization's solutions depending on their feasibilities. 

Having collected the data in \xuan{$S$} and \xuan{$\overline{S}$}, we use Algorithm \ref{alg:envelopefit} to fit the envelopes. With a specified number of envelopes, the algorithm first performs clustering with Gaussian Mixture Models (GMMs) to identify the center of clusters. Then, an optimization problem named Boundary\_fit is solved to fit the largest rectangular region $B$ around the centers excluding any points in \xuan{$\overline{S}_{i}$}. Multiple formulations can be used to achieve this. We used a formulation based on mixed-integer programming,
% other options include 
\ma{but other options exist}
\cite{deits2014convex}. 
% We then collect all the points inside $B$ and filter out the points with the probability that belongs to the cluster less than a threshold value $p_{th}$. 
\ma{We then collect all points in $B$ but remove those whose probability of belonging to the cluster is less than the threshold $p_{th}$.}
The exterior boundary formed by the remaining points gives \xuan{$\textbf{B}_{\mathcal{S}}$}.

\begin{algorithm}[t] 
\small 
\algsetup{linenosize=\small}
\caption{$\operatorname{EnvelopeFitting}$}
\label{alg:envelopefit}
\textbf{Input} Threshold probability $p_{th}$, Number of clusters $n$, 
% $S$, $\overline{S}$ from DataGeneration
DataGeneration's $S$, $\overline{S}$
% $S$, $S$ from DataGeneration
\begin{algorithmic}[1] \label{alg:envfit}
\STATE Initialize dictionary \xuan{$\textbf{B}_{\mathcal{S}}$}
\FOR{$k=1,...,n$}
\STATE Get cluster means \xuan{($p_{k}$, $f_{k}$)} = GMM(\xuan{$S$})
\STATE $B_{k}$ = Boundary\_fit(\xuan{($p_{k}$, $f_{k}$)}, \xuan{$\overline{S}$})
\STATE \xuan{$S_{k}$} = points in \xuan{$S$} that are in $B_{k}$ 
\STATE \xuan{$S^{o}_{k}$} = GMM\_filter($p>p_{th}$) 
\STATE \xuan{$\textbf{B}_{\mathcal{S}\_k}$} = get\_exterior\_boundary(\xuan{$S^{o}_{k}$})
\STATE Add \xuan{$\textbf{B}_{\mathcal{S}\_k}$} to \xuan{$\textbf{B}_{\mathcal{S}}$}
\ENDFOR
\RETURN \xuan{$\textbf{B}_{\mathcal{S}}$}
\end{algorithmic} 
\end{algorithm}

\subsubsection{Evolutionary Approach}
Contrasting to the fitting approach, a sampling-based evolutionary approach could, by nature, find bounds that are even tighter. 
Using a sufficient number of random terrain (input) and feasibility (output) pairs, a genetic algorithm (GA) can be tailored to solve a bilevel optimization that is indicative of the original problem at hand. While mostly following the conventional GA approach, we choose our chromosomes to be the lower and upper bounds of the envelope, while uniform crossover is done per lower/upper bound pair as opposed to per gene value. We design the fitness function to be representative of the bilevel optimization that occurs between the two stages of the original problem. To achieve this, we define two key metrics that help in finding better \xuan{envelopes}. We define $a$ to be the percentage of original infeasible values becoming feasible, and $b$ to be the percentage of original feasible values becoming infeasible.
% \begin{align*}
%     a = \text{\% of original infeasible values becoming feasible} \\
%     b = \text{\% of original feasible values becoming infeasible}
% \end{align*}
At each solution's fitness calculation, a random pair of terrain parameter and ground truth feasibility $z_g$ are selected from a pre-generated dataset built using the data generator. 
Then, the individual and the terrain parameter are used to find the feasibility $z_1$ in 
\ma{$P_1$.}
% the first stage of the two-stage optimization.
If $z_g$ is infeasible while $z_1$ is feasible, $a$ increases, whereas if $z_g$ is feasible but $z_1$ is infeasible, $b$ increases.
Per generation, each individual is tested against $K$ number of terrains from the training set and the average $a$ and $b$ are used in the calculation of the fitness function. 
While we try to minimize $a$, we keep $b$ below a certain threshold $\delta$. 
To ensure that the number of mutations also decreases over the generations and that the variance of the population's fitness decreases, the mutation rate is set to be a function of the generation number \xuan{decreasing} over time.

\section{Experiment} 
\subsection{Training}
\label{Sec:training_results}
\begin{figure}[!t]
		\centering
		\includegraphics[scale=0.24]{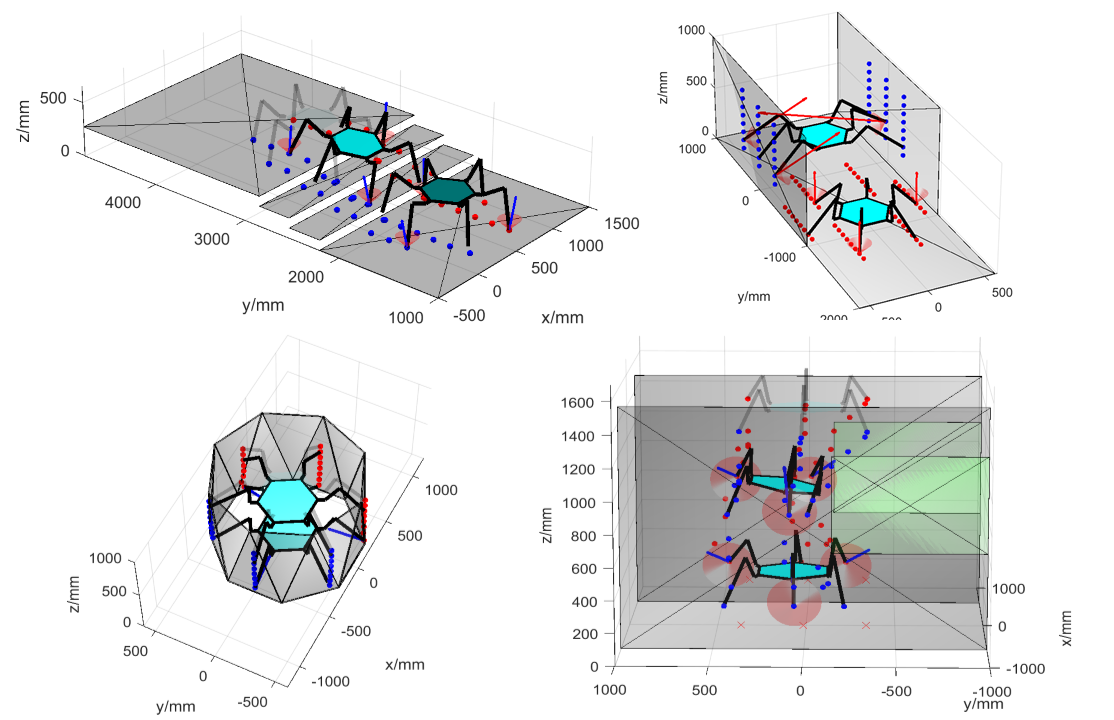}
		\caption{Motion plans generated by the proposed planner. Top left: the robot climbs stairs. Top right: the robot switches from walking to climbing. Bottom left: the robot climbs up inside a tube. Bottom right: the robot climbs up between two walls and avoid patches of low friction materials.}
		\label{Fig:All_planning_results}
\end{figure}

To validate our proposed algorithm,
% can fit McCormick envelopes that gives a 2-stage convex planner matching the original nonlinear optimization problem, 
we choose the random distribution $\theta$ to provide two different kinds of terrains---ground and wall---whose $(\textbf{p}, \textbf{f})$ are expected to show distinct distributions. For training, we provide one terrain mesh to each leg, while varying terrain position, orientation, and friction coefficient $\mu$ randomly.
% (a similar setup as shown in Fig. \ref{Fig:notations} with random meshes 1,2,3)
For ground data, we uniformly sample the angle of normal vectors within the \ang{30} $\textdegree$ region around the straight-up direction, while varying $\mu$ between [0.1, 0.8]. The wall data are collected to plan trajectories for the robot to climb up between two walls \cite{lin2018multi} with pure frictional contact. We vary the angles within the \ang{30} region around the nominal direction as shown in the top right of Fig. \ref{Fig:All_planning_results}, and vary $\mu$ between [0.1, 1.2]. We use Algorithm \ref{alg:datagen} to gather a set \ma{$S$} of 500 feasible points and 500 infeasible points for envelope fitting, and another set \xuan{$\overline{S}$} of the same number for validation. We set the number of envelopes to be 1. Both ground and wall envelopes are fit with 3 legs (two right, one left). Envelopes are also fitted with 5 legs on the wall. 
For validation, we separate the success rate into two categories showing respectively if our convex optimization can identify feasible solutions correctly and identify infeasible solutions correctly. We show the results of both stages. If \xuan{$\textbf{B}_{\mathcal{S}} \in S_{i}$}, $P_{1}$ should remain feasible if $P$ is feasible. If condition (2) in our theorem holds, $P_{1}$ should be infeasible if $P$ is infeasible. If condition (3) holds, the feasible solutions for $P_{1}$ should be projectable to make $P_{2}$ feasible, thus success rate for feasible solutions should not drop from $P_{1}$ to $P_{2}$. 
% The results are shown in Table \ref{tab:validation}.
\xuan{We use both convex and NLP methods mentioned in the previous section for $P_{2}$. For convex method, we use $\textbf{p}^{b}$ from $P_{1}$ to solve $\textbf{f}$ in $P_{2}$. The NLP method solves $\textbf{p}^{b}$ and $\textbf{f}$ together.}

\xuan{
% Human heuristics are used to create envelope parameters as baselines. 
\ma{Expert human heuristics are used to create envelope parameters as baselines. Bounds for $\textbf{p}^{b}$ are}
% We create the bounds for $\textbf{p}^{b}$ with 
measured outer boundaries of
% the workspace for each leg. 
\ma{each leg's workspace,}
% The bounds 
\ma{while that} for $\textbf{f}$ are 
% created out of 
\ma{from}
our understanding of force profiles. 
% The normal force bounds are $[0.0, 0.9]$ for climbing and $[0.0, 0.5]$ for walking, as in both cases large normal force are expected to prevent slipping. 
Normal force bounds are $[0.0, 0.9]$/$[0.0, 0.5]$ for climbing/walking, as large normal forces are expected to prevent slipping.
Shear force bounds are $[-0.3, 0.3]$/$[-0.15, 0.15]$ for climbing/walking. 
% In particular, f
For wall climbing the vertical shear force boundeds are $[0.0, 0.4]$ as they need to counteract the gravity. We perform hand-tuning to optimize the performance. The results are in Table \ref{tab:validation}.}

The results show that 
% we notice that using clustering, 
$P_{1}$ tends to match $P$ well for all test cases \ma{when using clustering}, as both success rates are close to 100\%. 
For convex projection, about 20\%$\sim$30\% of 
% 1st stage 
$P_1$'s solutions cannot be projected \xuan{for 3 leg case}. \xuan{This is due to a violation of condition (3)} indicating a single envelope may over-relax the original problem. For the 5 leg test, both stages perform well. Intuitively, since the problem dimension is higher, $\textbf{p}$ has more room to adjust at $P_2$. \xuan{NLP projections perform well, with the rate of projection close to $100\%$ and solving speed no more than a few hundred milliseconds.}
% the 2nd stage.

%(TODO:use two envelopes!)
Aside from training a single cluster, we tried to fit envelopes with combined ground and wall data. By giving $n=2$ in Algorithm \ref{alg:envelopefit}, two mutually exclusive envelopes representing the ground and the wall are identified.
This results in an MICP formulation with one binary variable $z \in \{0, 1\}$ per posture that switches between 2 modes as the robot traverses from one type of terrain to the other.

\begin{table}[t]
\caption{Validation results for trained envelopes}
% \begin{adjustbox}{width=0.48\textwidth}
% \small
\begin{center}
\scalebox{0.7}{
\begin{tabular}{cc|cc|c|c}
\hline
\multicolumn{2}{c|}{\multirow{2}{*}{Problem}}                                       & \multicolumn{2}{c|}{$P_1$}                                                                                                   & $P_1 \rightarrow P_2$                                                      & $P_1 \rightarrow P_2$ (NLP)                                     \\ \cline{3-6} 
\multicolumn{2}{c|}{}                                                               & \begin{tabular}[c]{@{}c@{}}correct\\ feasible\end{tabular} & \begin{tabular}[c]{@{}c@{}}correct\\ infeasible\end{tabular} & \begin{tabular}[c]{@{}c@{}}correct\\ feasible$*$\end{tabular} & \begin{tabular}[c]{@{}c@{}}correct\\ feasible\end{tabular} \\ \hline
\multirow{3}{*}{\begin{tabular}[c]{@{}c@{}}3 leg\\ ground\end{tabular}} & Cluster   & 98.16\%                                                    & 77.78\%                                                      & 83.54\%                                                    & 98.16\%                                                    \\ \cline{2-6} 
                                                                        & GA        & 70.08\%                                                    & 91.24\%                                                      & 37.40\%                                                    & 70.08\%                                                    \\ \cline{2-6} 
                                                                        & Heuristic & 99.77\%                                                    & 50.00\%                                                      & 84.91\%                                                    & 99.77\%                                                    \\ \hline
\multirow{3}{*}{\begin{tabular}[c]{@{}c@{}}3 leg\\ wall\end{tabular}}   & Cluster   & 93.62\%                                                    & 80.00\%                                                      & 63.32\%                                                    & 93.62\%                                                    \\ \cline{2-6} 
                                                                        & GA        & 77.85\%                                                    & 97.72\%                                                      & 67.07\%                                                    & 77.85\%                                                    \\ \cline{2-6} 
                                                                        & Heuristic & 78.72\%                                                    & 51.33\%                                                      & 60.75\%                                                    & 78.72\%                                                    \\ \hline
\multirow{3}{*}{\begin{tabular}[c]{@{}c@{}}5 leg\\ wall\end{tabular}}   & Cluster   & 95.27\%                                                    & 79.31\%                                                      & 90.41\%                                                    & 93.49\%                                                    \\ \cline{2-6} 
                                                                        & GA        & 68.21\%                                                    & 98.68\%                                                      & 62.70\%                                                    & 68.21\%                                                    \\ \cline{2-6} 
                                                                        & Heuristic & 92.90\%                                                    & 72.41\%                                                      & 88.10\%                                                    & 92.25\%                                                    \\ \hline
\end{tabular}
}
\end{center}
\vspace{-2ex}
\begin{tablenotes}\footnotesize
\item[*] Correct infeasible results for $P_{2}$ are identical to $P_{1}$ thus omitted.
\end{tablenotes}
% \end{adjustbox}
\label{tab:validation}
\end{table}

\begin{figure}[!b]
		\centering
		\includegraphics[width=0.45\textwidth]{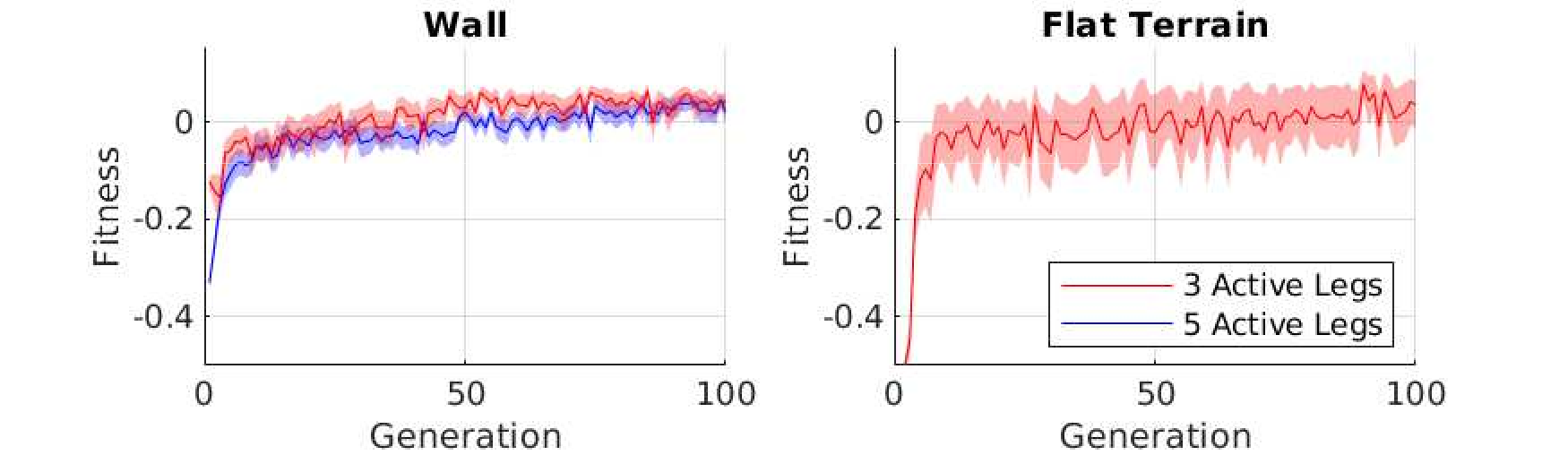}
		\caption {The average fitness of the population is shown with its variance. Certain individuals maximize the fitness early on in the generation.}
		\label{Fig:fitnessvsgen}
\end{figure}

Contrary to the clustering approach, the GA approach shows lopsided results. This could be an artifact of suboptimal initial solutions and a lack of generations as seen by the variance in Fig. \ref{Fig:fitnessvsgen}. However, the extremely high infeasibility detection suggests that possibly a combined approach between a structured clustering and a conventional learning based approach could achieve higher accuracies.

\ma{The results suggest that our methodological approach based on data immediately provides comparable results to heuristics that require expert knowledge. }

% We first try walking on the ground with 3 leg support. The results show this problem is largely decoupled, which should not be a surprise.

% We then try climbing between 2 walls with 3 leg support. The results show this problem is highly coupled. We supply a mathematical process demonstrating how to get similar bounds mathematically using static indeterminacy analysis.

% Since a single McCormick envelope still have about 20\% chance to fail we tried using 2 envelopes, resulting in training MICP.

% Finally, we try climbing between 2 walls with 5 leg support, the same experiment as we did in previous paper \cite{lin2019optimization}. The results show this problem is largely decoupled, which may be less intuitive. However, if we recognize that only 3 legs are necessary in supporting the robot and any extra leg is redundant, 

% This provides an additional justification for our decoupled approach in previous paper \cite{lin2019optimization}. It is handy that GA can find all those heuristics for us.

% When training, maybe use heuristics to initialize the parameters, and mention we can get upper/lower bounds.

% To compare with the heuristics that we came up with
% Justify why GA is even chosen as the first algorithm

% \begin{figure}[!t]
% 		\centering
% 		\includegraphics[scale=0.4]{Figures/Learned_envelope.eps}
% 		\caption{}
% 		\label{Fig:Fitted_envelope}
% \end{figure}

% \begin{figure}[!t]
% 		\centering
% 		\includegraphics[scale=0.4]{Figures/Learned_envelope_2modes.eps}
% 		\caption{}
% 		\label{Fig:Fitted_envelope}
% \end{figure}

\subsection{Planning}
\label{Sec:planning_results}

% say why we choose this experiment, because it is complicated and the decompose approach will fail.

% To test the performance of the trained envelopes, they are applied to the 2-stage convex planner. 
We use the learned envelopes in our 2-stage planner, and plan the motion for a 24 DoF hexapod.
%robot's optimization-based motion planner, which requires a high-dimensional envelope for the bilinear constraints. 
Trajectories to traverse in multiple terrains are found, including walking on flat ground with stairs, climbing between two flat vertical walls of varying friction, and climbing inside a tube. 
% We then use the planner to solve the trajectories for a 24-DoF hexapod robot in multiple environments including walking on flat ground with stairs (if slope should be here is to be determined), climbing between two flat vertical walls of varying coefficient of friction, and climbing inside a tube. 
% The unique capability of the coupled approach is compared against the decoupled approaches in \cite{lin2019optimization}. 
The coupling achieved in the planner is compared against the decoupled approaches \cite{lin2019optimization}.
% The planner is then tested on the real hardware. All results are included in the attached video.
% If one robot leg needs to select among multiple contact planes, we implement an MICP formulation to choose contact planes for each leg.
The planner is verified on hardware. All results are included in the accompanying video \cite{youtube_video}. 

\subsubsection{Walking on Ground}
The proposed planner generates trajectories for walking on a flat ground with stairs (Fig. \ref{Fig:All_planning_results} top left). 
This shows the feasibility of the approach on a simple environment because the terrains are effectively flat. 
% As we have shown in the last section, the problem is largely decoupled, meaning a simple 2-step approach that first plans footstep position then plans contact forces will work well. 
% The hardware implementation is provided in the attached video.

\subsubsection{Climbing between Walls}
% We implemented the approach on the physical platform, which receives position commands as opposed to force commands. This requires an additional layer of optimization to convert force commands to position commands.

% To further assess the planner, trajectories to let the robot climb up between two walls are planned for. 
This problem is first studied by \cite{lin2018multi}, which uses a stiffness based approach allowing the robot to brace between two walls and climb. A 2-stage decoupled approach is then used \cite{lin2019optimization} to plan the climbing motion, where it plans the position $\textbf{p}$'s without any knowledge of force.
%Climbing between two flat walls are initially done as seen in Fig. \ref{Fig:All_planning_results} (top right), with each wall being composed of one flat region. 
To test the proposed coupled planner, we plan the trajectory for the robot to climb on the walls with patches of zero friction $\mu=0$ (green regions in Fig. \ref{Fig:All_planning_results}, bottom right) and $\mu=1$ in other areas. 
% The 1st stage planner needs to place toes outside the region. 
% If not, the 2nd stage planner will be infeasible. 
\ma{$P_1$ will have to place the toes outside the region to avoid $P_2$ being infeasible.}
For this, decoupled planner \cite{lin2019optimization} would fail.
We also plan a climbing motion inside a tube consisting of meshes at different angles (Fig. \ref{Fig:All_planning_results}, bottom left), to test the adaptability to irregular walls. 
In both cases, the planer solves feasible motion plan on hardware.

\subsubsection{Automatic switching motion plan from ground to wall}
\ma{We demonstrate that with the learned 2 mode formulation presented in Section \ref{Sec:training_results}, 
% the motion plan can generate 
trajectories with contact forces that automatically switch between 
% walking mode and climbing mode
walking and climbing can be generated as seen in Fig. \ref{Fig:All_planning_results} top right.} 
% As shown in the up right figure in Fig. \ref{Fig:All_planning_results}, we plan a trajectory that includes a switching point from ground to wall. 
% In addition to
\ma{Beyond} finding a feasible trajectory, 
% when the robot is on the ground, the planner finds $z=0$, indicating that it interprets the current motion as walking. When the robot is on the wall, the planner finds $z=1$, indicating that it interprets the current motion as climbing. 
\ma{the planner finds $z=0$/$z=1$ when the robot is on the ground/wall, indicating that it interprets the current motion as walking/climbing.}
This result shows the interpretability of the proposed planner. If we divide the variable space into smaller pieces and assign an integer variable for each piece, the information found by the planner can quickly get submerged by the large number of possible combinations of integer variables. Instead, our planner gives interpretable information that humans can process.

\begin{table}[] 
\caption{Solving time for vertical two flat wall climbing}
\begin{threeparttable}[t]
\begin{center}
\resizebox{1\linewidth}{!}{
\begin{tabular}{ccccc}
\hline
\textbf{Test Name}    & \textbf{Rounds $M$} & \textbf{Variables} & \textbf{Constraint} & \textbf{Solve Time [s]\tnote{*}} \\ \hline \hline
2-stage Convex   & 8               & 4128              & 1176                 & 0.16                   \\ \hline \hline
MICP without Envelope   & 4               & 16608 (276 binary)              & 3360                 & 91                   \\ \hline
MICP with Envelope    & 4               & 16608 (276 binary)             & 4008                 & 33                   \\ \hline \hline
MICP without Envelope   & 8               & 33216 (552 binary)             & 6720                 & $>$ 1,000              \\ \hline
MICP with Envelope    & 8               & 33216 (552 binary)              & 8016                 & 152                 \\ \hline
\end{tabular}}
\end{center}
\begin{tablenotes}\footnotesize
\item[*] Data taken on an Intel i5-6260U 1.80GHz machine with Gurobi \cite{optimization2014inc}
\end{tablenotes}
\end{threeparttable}
\label{tbl:vwall_data}
\end{table}

\subsubsection{Solving Time}

We benchmark the solving time on the problem of climbing between two flat walls with uniform $\mu$. 
Table \ref{tbl:vwall_data} row 1 shows the solving time for the proposed two-stage convex planner. 
The convex solver generates trajectories in hundreds of milliseconds. 
% Because the optimization algorithm is convex, it is fast. 
% With a similar problem size, it runs 100 times faster than an NLP solver (refer to Table I in \cite{winkler2018gait}). 
Comparison with similar works using NLPs suggests a possible speed up of around 100 times \cite{winkler2018gait}. 
% This justifies the original motivation---a multi-staged convex algorithm tends to run faster than a single stage NLP solver. 
This justifies the motivation behind using a multi-staged convex algorithm. 
% We also show that the envelope that we designed can be used to speed up the mixed-integer convex programming. 
\ma{The designed envelope can also be used to speed up MICP.}
% We implement the accurate MICP formulation \cite{dai2019global} to solve the problem of climbing between two flat walls, and compare it against the identical formulation but add our learned envelope as an additional constraint. 
\ma{We compare the accurate MICP formulation with and without our learned envelope as an additional constraint for the problem of climbing between two flat walls.}
A typical MICP solver deals with integer variables through a branch and bound algorithm \cite{williams2013model} that expands nodes on each binary variable. 
Since 
% the formulation in 
\cite{dai2019global} does not utilize the problem-specific knowledge, 
% the solver will expand nodes inside the regions that are known to be infeasible by the problem knowledge, wasting a large amount of time.
\ma{the solver wastes time expanding nodes inside regions that we already know are infeasible.}
% With the learned envelopes, 
Our approach reduces the solving time, 
% while has
\ma{and has}
minimum impact on the solutions (Table \ref{tbl:vwall_data}, row 2-4). 

% The inherent nature of footstep selection, which necessitates mixed-integer programming, also extends solve times. 
% However, this can also be additionally reduced through our learned envelopes, which when compared with similar approaches that try to find better relaxations albeit with a lack of domain knowledge, tends to be faster. 

% If the robot has to select footstep among multiple contact regions, the mixed-integer programming comes into play, which slows down the solver. 
% We compare the MICP solving speed with our learned envelope approach against the evenly divided approach in \cite{dai2019global}, which does not utilize the problem specific knowledge. The solving time is compared for climbing between two flat walls. We show that the learned envelope approach significantly cuts down the solving time, while has minimum impact on the solution found (Table \ref{tbl:vwall_data}, row 2-4). 

\section{Conclusion, Discussion and Future Work} 
\label{Sec:conclusion}
% Two things are critical for implement of more complicated MICP:
% 1. gradient
% 2. cannot directly sample MICP, need to go through the search tree

% GA is difficult to guarantee working

% Interesting to see if algorithm can "bifurcate" a single envelope into multiple envelopes if one seems to be not enough.

% Is there a different dimensional representation of the envelope that might be more useful? % (Unfortunately yes there seems to be...) - ??

% Our paper proposed 
We propose
a 2-stage motion planner based on convex optimization, with the inter-stage coupling formulated as McCormick envelopes learned from data. We 
% perform
\ma{performed}
learning through clustering and GA approaches and validated against labeled data. 
The results show that a smaller number of integer variables and envelopes tailored to the type of problem
% can benefit interpretability and solving speed. 
\ma{can reduce solve time and help interpret the outcome.}
We also demonstrated the planner on the hardware.

% When one envelope becomes insufficient, algorithms may be designed to cut the region into smaller pieces, hence finer relaxations.
\ma{Finer relaxations with smaller envelopes could be possible if one envelope is insufficient.}
Learning an objective function that guides the 1st stage solution to a projectable point is another future work. Since 
% our algorithm requires 
we require
exploring the solution set,
% which may be done efficiently on the hardware,
\ma{which could be efficient on hardware,}
%   direct 
training on hardware is of interest. 
% These results, along with the results from \cite{cauligi2020learning}, suggest that it may be helpful to classify the input data before sending to MICP or MINLP solvers. 
% While focusing on a specific problem, our algorithm may be general for more complicated optimization problems with more than 2 stages, and a larger class of nonlinear constraints. 
While we focused on a specific problem, this work may be generalizable for more complicated problems with multiple stages, and a larger class of nonlinear constraints. 

%\section*{APPENDIX}
%\input{sect_appendix}

% % Appendixes should appear before the acknowledgment.

{
\bibliographystyle{IEEEtran}
\bibliography{references}
}

\end{document}